\documentclass[twocolumn,letterpaper]{IEEEAerospaceCLS}  %
\pdfoutput=1 
\usepackage[]{graphicx}
\newcommand{\ignore}[1]{} 
\usepackage[hyphens]{url}
\usepackage{hyperref} 
\hypersetup{hidelinks}

\usepackage{tabularx}
\newcolumntype{Y}{>{\centering\arraybackslash}X}

\usepackage{booktabs}

\usepackage{placeins}

\usepackage{siunitx} 
 
\usepackage{listings}

\usepackage{orcidlink}

\usepackage[autostyle]{csquotes}
\MakeOuterQuote{"}

\usepackage{float}

\usepackage{tabularx} 

\begin{document}
\title{A Comparison of Deep Learning Architectures\\for Spacecraft Anomaly Detection}

\author{%
\href{https://orcid.org/0000-0002-8198-7892}{Daniel Lakey}~\orcidlink{0000-0002-8198-7892}\\ 
IU International University of Applied Sciences\\
daniel.lakey@iu-study.org
\and 
\href{https://orcid.org/0000-0002-9462-8610}{Tim Schlippe}~\orcidlink{0000-0002-9462-8610}\\
IU International University of Applied Sciences\\
tim.schlippe@iu.org
\thanks{\footnotesize \copyright 2024 IEEE. Personal use of this material is permitted. Permission from IEEE must be obtained for all other uses, in any current or future media, including reprinting/republishing this material for advertising or promotional purposes, creating new
collective works, for resale or redistribution to servers or lists, or reuse of any copyrighted
component of this work in other works.} 
}

\maketitle

\thispagestyle{plain}
\pagestyle{plain}

\maketitle

\thispagestyle{plain}
\pagestyle{plain}

\begin{abstract}
Spacecraft operations are highly critical, demanding impeccable reliability and safety. Ensuring the optimal performance of a spacecraft requires the early detection and mitigation of anomalies, which could otherwise result in unit or mission failures. With the advent of deep learning, a surge of interest has been seen in leveraging these sophisticated algorithms for anomaly detection in space operations. Our study aims to compare the efficacy of various deep learning architectures in detecting anomalies in spacecraft data.
The deep learning models under investigation include Convolutional Neural Networks (CNNs), Recurrent Neural Networks (RNNs), Long Short-Term Memory (LSTM) networks, and Transformer-based architectures. Each of these models was trained and validated using a comprehensive dataset sourced from multiple spacecraft missions, encompassing diverse operational scenarios and anomaly types. 
We also present a novel approach to the rapid assignment of spacecraft telemetry data sets to discrete clusters, based on the statistical characteristics of the signal. This clustering allows us to compare different deep learning architectures to different types of data signal behaviour.
Initial results indicate that while CNNs excel in identifying spatial patterns and may be effective for some classes of spacecraft data, LSTMs and RNNs show a marked proficiency in capturing temporal anomalies seen in time-series spacecraft telemetry. The Transformer-based architectures, given their ability to focus on both local and global contexts, have showcased promising results, especially in scenarios where anomalies are subtle and span over longer durations.
Additionally, considerations such as computational efficiency, ease of deployment, and real-time processing capabilities were evaluated. While CNNs and LSTMs demonstrated a balance between accuracy and computational demands, Transformer architectures, though highly accurate, require significant computational resources.
In conclusion, the choice of deep learning architecture for spacecraft anomaly detection is highly contingent on the nature of the data, the type of anomalies, and operational constraints. This comparative study provides a foundation for space agencies and researchers to make informed decisions in the integration of deep learning techniques for ensuring spacecraft safety and reliability.
\end{abstract}

\tableofcontents

\setcounter{footnote}{0}

%%%%%%%%%%%%%%%%%%%%%%%%%%%%%%%%%%%%%%
\section{Introduction}
%%%%%%%%%%%%%%%%%%%%%%%%%%%%%%%%%%%%%%
The field of space exploration has had significant advancements in recent decades, characterised by the increasing sophistication of spacecraft and the expanding complexity of missions. As mankind expands its presence in outer space, the importance of precise and dependable data from spacecraft systems has become of utmost significance. Time series data, which refers to a sequential arrangement of data points organised in chronological order, holds major significance in the domain of spacecraft telemetry. Spacecraft systems are reflected by telemetry data, which provides information on their state, health, and performance. This data allows for the analysis of both regular and potentially abnormal operations~\cite{zacchei.2003}.

Anomalies observed in spacecraft telemetry data are unanticipated occurrences that pose potential risks, as they depart significantly from the predicted operational patterns of the system. The quick detection and identification of these abnormalities is of paramount significance in order to avert catastrophic failures, limit risks, and guarantee the durability of space missions. According to \cite{Guan.2022}, the prompt identification and effective detection of these anomalies by operational engineers play a crucial role in enhancing efficiency, minimising expenses, and enhancing safety. As the complexity of spacecraft continues to advance, there is a corresponding growth in the variety of telemetry parameters associated with them. The utilisation of conventional, manual or simple "out-of-limits" techniques are becoming ever more difficult for the purpose of identifying anomalies~\cite{Heras.2014}.

In recent years, there has been considerable focus on the advancement of anomaly detection techniques for satellite telemetry data. Numerous advanced algorithms and strategies have been proposed by prominent organisations such as NASA~\cite{Hundman.2018}, ESA~\cite{Heras.2014}, and CNES~\cite{Pilastre.2020} to tackle this task. Every approach possesses its own set of advantages and disadvantages. There is a clear trend towards deep learning approaches over statistical methods due to their ability to synthesise the complex multivariate temporally-connected data inherent to spacecraft telemetry~\cite{Schmidl.2022}. The objective of this paper is to investigate and assess different methodologies for anomaly identification in order to determine the most optimal and efficient approach for analysing spacecraft telemetry.

Our work pioneers several notable contributions to the domain of spacecraft anomaly detection, presenting advancements that enhance the understanding of deep learning in this field. Firstly, it unfolds a comprehensive side-by-side comparison of multiple deep learning model architectures, shedding light on their effectiveness in detecting anomalies in spacecraft telemetry. This comparison is distinctively valuable as it incorporates models that, to our knowledge, have not been previously applied to spacecraft anomalies, thereby opening new avenues for exploration and implementation. Secondly, we introduce an innovative unsupervised mechanism to cluster spacecraft telemetry into like-types, using statistical methods, which allows for a more granular and nuanced understanding of telemetry data. Thirdly, our study unveils insights into the comparative performance of different deep learning models across the identified clusters, providing insights for selecting the most suitable model based on the specific type of telemetry data. These diverse contributions collectively elevate the current state of research in spacecraft anomaly detection, offering robust and refined tools and methodologies for practical applications and future explorations.

%%%%%%%%%%%%%%%%%%%%%%%%%%%%%%%%%%%%%%
\subsection{Nomenclature}
%%%%%%%%%%%%%%%%%%%%%%%%%%%%%%%%%%%%%%
The following terms are used in this work.  

Telemetry Channel; A specific pathway or conduit used for transmitting telemetry data~\cite{Chapman_Critchlow_Mann_1963}, for example from a specific sensor on the spacecraft. A telemetry channel consists of one or more parameters.

Parameter; A measurement within a telemetry channel. This may be an analogue reading such as temperature or current, a discrete numerical value or a binary status. A telemetry time series is made up of many samples of a number of parameters representing a number of telemetry channels.

Dataset; A collection of data points or individual pieces of information, organised usually in tabular form, where rows represent individual records and columns represent attributes or variables of the data. The dataset used in our study is from~\cite{Hundman.2018} and contains data for 82 telemetry channels.

Cluster; A grouping of data points or items in a dataset that share similar characteristics or properties, typically identified through various methods of cluster analysis, allowing for the study of relationships and patterns within the data~\cite{Omran.2007}.

%%%%%%%%%%%%%%%%%%%%%%%%%%%%%%%%%%%%%%
\section{Related Work}
%%%%%%%%%%%%%%%%%%%%%%%%%%%%%%%%%%%%%%
This section will describe related work investigated as part of our study, focusing especially on popular deep learning architectures, many of which have not been applied to the problem of anomaly detection in spacecraft telemetry data.

%%%%%%%%%%%%%%%%%%%%%%%%%%%%%%%%%%%%%%
\subsection{Data for Spacecraft Anomaly Detection}
%%%%%%%%%%%%%%%%%%%%%%%%%%%%%%%%%%%%%%
Modern spacecraft have many thousands of telemetry channels~\cite{Baireddy.2021}, and this "huge"~\cite{yairi2014evaluation} amount of data is more than can be monitored by human operators. Within these channels, actual instances of anomalies are rare. By design a spacecraft is a robust machine, fault tolerant and extensively tested to ensure that anomalies do not occur~\cite{Fortescue_Swinerd_Stark_2011}. For example, a study of seven different spacecraft over more than a decade yielded fewer than 200 critical anomalies~\cite{lutz2004empirical}.

Spacecraft Anomaly Detection is a particularly challenging field due to the sparsity of publicly available datasets for training. Indeed, of all the studies listed in our work, only \cite{Hundman.2018} make the data available, and even then with implementation-specific details hidden through scaling and normalisation. This has led to their dataset becoming a benchmark for further studies, such as \cite{Baireddy.2021} and consequently we used it in our experiments.

%\subsubsection{SMAP/MSL Dataset}
The dataset provided by \cite{Hundman.2018} comprises of 82 telemetry channels taken from the Soil Moisture Active Passive (SMAP)~\cite{smap} spacecraft and "Curiosity" Mars Science Laboratory (MSL)~\cite{msl} spacecraft. In Section~\ref{lab:data}, we will describe this dataset in the context of our experimental setup.

%%%%%%%%%%%%%%%%%%%%%%%%%%%%%%%%%%%%%%
\subsection{Approaches for Spacecraft Anomaly Detection}
%%%%%%%%%%%%%%%%%%%%%%%%%%%%%%%%%%%%%%

\begin{figure}
    \centering
    \includegraphics[width=1\linewidth]{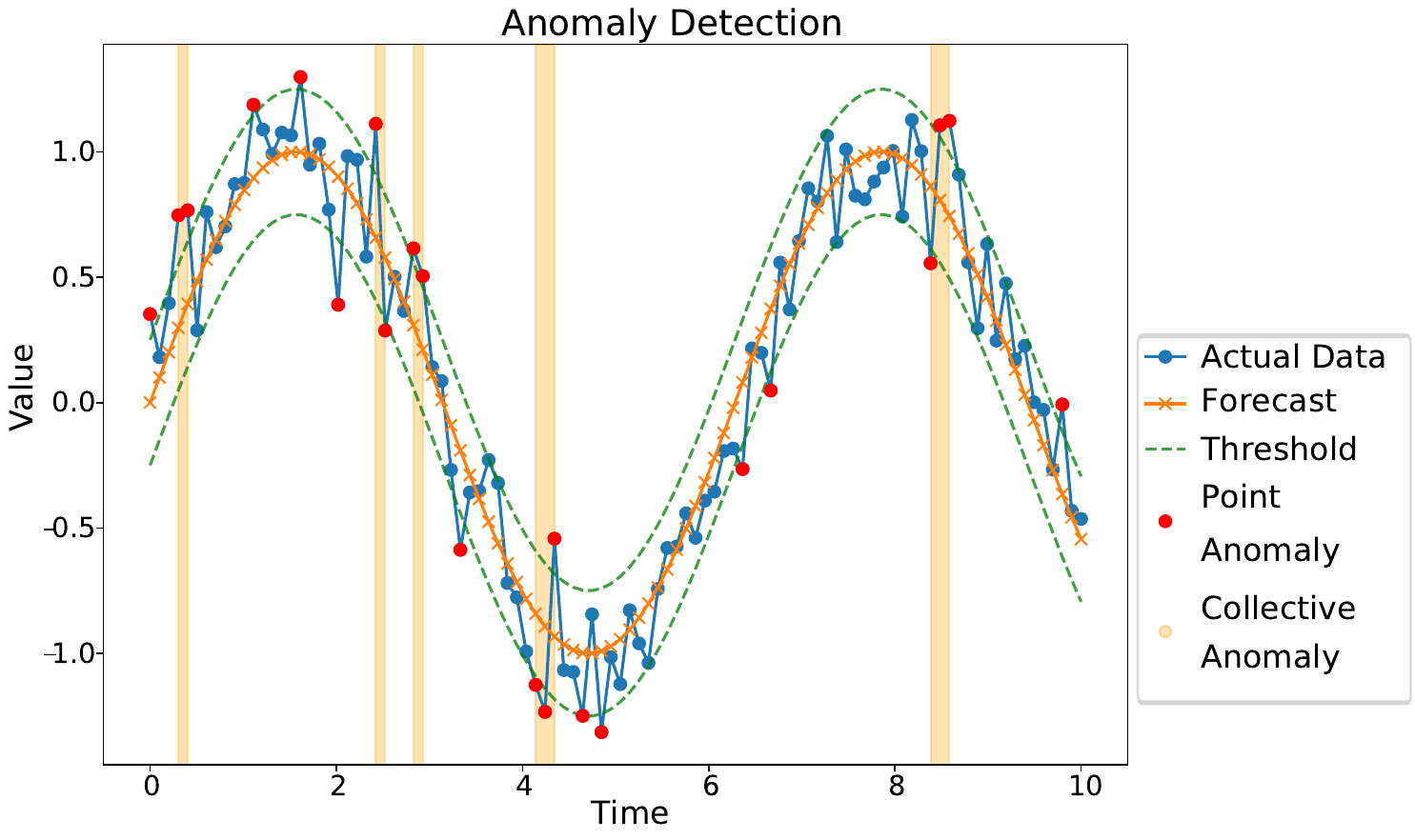}
    \caption{\bf{Anomaly Detection with Forecasting and Thresholding}}
    \label{fig:anomaly_detection}
\end{figure}

A typical approach, for example followed by \cite{Heras.2014},~\cite{Hundman.2018},~\cite{Pilastre.2020}, and~\cite{Baireddy.2021}, is the use of deep learning models to perform regression-based forecasting on a time series and identify anomalies by comparing predictions to real values received from the spacecraft. The central concept is "\textit{to reconstruct the telemetry sequence based on training data, and anomalies are identified if the reconstruction errors exceed a given threshold.}"~\cite{He.2022}, as illustrated in Figure~\ref{fig:anomaly_detection}. "\textit{The idea is to use past telemetry describing normal spacecraft behaviour in order to learn a reference model to which can be compared most recent data in order to detect potential anomalies.}"~\cite{Pilastre.2020}. Multivariate models are used to capture spatial and temporal linkages between separate telemetry channels~\cite{chakraborty1992forecasting}. 

Whilst effective, it relies on the selection of some threshold value beyond which the construction error is considered anomalous. \cite{Hundman.2018} propose "Telemanom", an "\textit{unsupervised and nonparametric anomaly thresholding approach}" where the anomaly detector dynamically learns the error value corresponding to the anomaly for a particular time series. They report excellent F1 scores for the anomaly detection, as synthesised in Table~\ref{tab:telemanom_perf}, which we use as our baseline.

\begin{table}[!h]
    \centering
    \caption{\bf{Telemanom F1 Scores}}
    \label{tab:telemanom_perf}
    \begin{tabular}{c|c|c}
        \textbf{SMAP} & \textbf{MSL} & \textbf{Total}\\ \hline
        85.5\% & 79.3\% & 83.6\%
    \end{tabular}
\end{table}

There exist many types of architecture for deep learning, many of which have been tuned specifically to time series prediction, for example~\cite{Walther.2023}. We selected six state-of-the-art families of architecture for further investigation. Additionally, we investigated two hybrid architectures comprised of a combination of two or more model types as suggested by~\cite{Han.2023}.

\subsection{Chosen Deep Learning Architectures for our Study}\label{lab:archs}
The following sections briefly review the chosen deep learning architectures used this our study. In particular, it is noted whether previous work has tried these in the domain of spacecraft anomaly detection and which are novel to this task.

\subsubsection{Multilevel Wavelet Decomposition}\label{lab:mwdn}
~\cite{mWDN} introduced their Multilevel Wavelet Decomposition Network (mWDN) for anomaly detection. mWDN leverages the benefits of wavelet transformation in conjunction with a deep learning model to analyse time series data, with a specific emphasis on interpretability. Wavelet transformation is a powerful mathematical tool often used for analysing different frequency components in time series data, which makes it highly suitable for anomaly detection in varied applications such as high-frequency signals~\cite{Michau2022} and power converters~\cite{Ye.2022}. To the best of our knowledge, an mWDN has yet to be applied to spacecraft telemetry anomaly detection.

\subsubsection{Multi-Layer Perceptron (MLP)}\label{lab:mlp}
The gMLP~\cite{liu2021pay}, or gated Multi-Layer Perceptron, is a type of artificial neural network model designed to have performance competitive to Transformer models but with a more straightforward architecture. It relies more on feedforward layers and less on attention mechanisms. A gMLP utilises Spatial Gating Units (SGU), a central component that enables information exchange between different positions in the sequence, allowing the model to capture dependencies between different parts of the input. MLPs have been used for anomaly detection in fields as varied as water treatment~\cite{RAMANMR2020100393} to rogue trading~\cite{Hedström_Wang_2021}, as well as spacecraft anomaly detection~\cite{bernal2021machine}.

\subsubsection{Transformer}\label{lab:transformer}
Transformers, originally proposed in \cite{vaswani2023attention}, are a type of neural network architecture that have become the foundation for most state-of-the-art models in natural language processing, and they are increasingly being used in various domains like time series analysis and image processing. Transformers use a mechanism called self-attention that allows each element in the input sequence to consider other elements in the sequence when producing its output, weighting each one differently depending on the learnt relationships. Transformers are the subject of much active research into anomaly detection, such as \cite{KIM2023105964}, \cite{jeong2023anomalybert} and \cite{xu2022anomaly}. The implementation of the Transformer   architecture investigated in our study is TimeSeriesTransformer (TST)~\cite{tst}, which tunes the architecture specifically for multivariate time series data, of which spacecraft telemetry is an extreme example owing to the potentially very large number of parameters to consider~\cite{Meng2020SpacecraftAD}.

\subsubsection{Convolutional Neural Network (CNN)}\label{lab:cnn}
Convolutional Neural Networks (CNNs) are a class of deep learning models primarily developed for analysing visual imagery, renowned for their ability to learn hierarchical features from input data~\cite{LeCun1989HandwrittenDR}. In the context of spacecraft anomaly detection, CNNs can be utilised to process multivariate time series data generated by spacecraft sensors~\cite{Yue.2022}, enabling the identification of anomalous patterns or events indicative of potential faults, malfunctions, or other abnormalities in spacecraft systems~\cite{Tennberg_Ekeroot_2021}. By learning both spatial and temporal features in the data, CNNs can aid in early and accurate detection of anomalies~\cite{FANAEET2016130}. As a popular architecture in deep learning, there are many implementations of interest. Four are selected here, including the "classics" ResNet~\cite{he2015deep,wang2016time} and Fully Convolutional Network (FCN)~\cite{wang2016time}, in addition to some implementations specifically tailored to time series: XceptionTime~\cite{rahimian2019xceptiontime} and InceptionTime~\cite{Ismail_Fawaz_2020}. For InceptionTime, two implementations from \verb|tsai| we chose : MultiInceptionTimePlus and InceptionTimeXLPlus. The former is an ensemble method with multiple internal models, whereas the latter contains a large number of parameters.

\subsubsection{Recurrent Neural Network (RNN)}\label{lab:rnn}
Recurrent Neural Networks (RNNs) are a category of neural networks specialised for processing sequential data, enabling the modeling of temporal features within the sequences. Our study includes two RNN variants. Long Short-Term Memory (LSTM)~\cite{lstm} units are a variant of RNNs designed to mitigate the vanishing and exploding gradient problems inherent in basic RNNs, allowing them to learn long-range behaviours within the data. The model used in our baseline study~\cite{Hundman.2018} is LSTM-based. Gated Recurrent Units (GRU)~\cite{gru} are another variant of RNNs, similar to LSTMs but with a simpler structure, designed to capture dependencies for sequences of varied lengths. GRUs have been proven to perform comparably to LSTMs on certain tasks~\cite{cahuantzi2023comparison} but with reduced computational requirements, offering an efficient alternative for sequence modeling. This may make them of particular use in spacecraft anomaly detection~\cite{Xiang.2021}, where the cost of training the more complex models may be prohibitive. 

\subsubsection{Hybrid Models}\label{lab:hybrid}
To leverage the advantages of different deep learning architectures, many previous studies have considered hybrid models~\cite{Lin.2020,Han.2023}. LSTM/Transformer models are quite popular, for example \cite{Andayani.2022} and \cite{zeng2020leveraging}, and in our study we include two such models in the suite of tested architectures, TransformerLSTM~\cite{Urazalinov_2023},~\cite{tsai} and LSTMAttention~\cite{tst},~\cite{vaswani2023attention},~\cite{tsai}. Other studies~\cite{Sanderson.2022} have considered a hybrid FCN/Transfomer model, combing the spatial learning abilities of CNNs with the sequence learning of Transformers, therefore we include LSTM\_FCN~\cite{tsai} in our test set. Other studies such as ~\cite{math11030676} go further still andcombining CNN, RNN and Transformer architectures in one model.

 We are unaware of the use of hybrid deep learning models for spacecraft anomaly detection, although there have been studies in the area of hybrid machine-learning such as \cite{He.2022} and \cite{XU2023281}.

%%%%%%%%%%%%%%%%%%%%%%%%%%%%%%%%%%%%%%
\section{Experimental Setup}
%%%%%%%%%%%%%%%%%%%%%%%%%%%%%%%%%%%%%%
This section describes the details of the experimental  setup used for the comparison of deep learning approaches.

\subsection{Implementation}
 We employed the Telemanom~\cite{Hundman.2018} anomaly detection framework for conducting experiments on spacecraft anomaly detection. The original implementation of Telemanom uses a \verb|Tensorflow|-based LSTM model as a default, designed to recognize anomalous patterns in time series data relevant to spacecraft telemetry. We will replace this LSTM with alternative architectures as described in Section \ref{lab:archs}.

We retain the dynamic thresholding and anomaly detection algorithms of Telemanom whilst replacing the time-series forecasting models with a variety of new architectures. Thus, we can clearly demonstrate the differences due to the architecture alone. To this end, the default LSTM was replaced with various models provided by the \verb|tsai| library~\cite{tsai}, a PyTorch and \verb|fastai|-based collection of time-series deep learning architectures~\cite{Howard_2020}. The \verb|tsai| library implements a wide selection of state-of-the-art models optimised for time-series data. In order to keep the model code generic and easily fit to a variety of different model architectures, the \verb|tsai| "Plus" implementations of the above architectures were selected due to their common interfaces. Detailed documentation regarding the particulars of implementation can be found in \cite{tsai}.

Following the approach taken in \cite{Hundman.2018}, one model is trained per telemetry channel. Our study compares thirteen different architectures, leading to \(82 \times 13 = 1,066\) trained models overall. 

The models were trained utilizing the "fit one cycle" method~\cite{smith2018disciplined}, a technique noted for its efficacy in training deep learning models efficiently and reliably. The experiment endeavoured to keep the setup fair and comparable; thus, hyperparameter tuning was predominantly confined to ensuring that the RNN-based architectures possessed at least equivalent depth to the default LSTM implemented in Telemanom. Apart from this modification, we retained the default hyperparameters provided by the \verb|tsai| and \verb|fastai| models to maintain the integrity of the comparative analysis, on the basis that the defaults are anyway sensible~\cite{Howard.2021}. Furthermore, due to the large number of trained models, hyperparameter tuning was infeasible in any case.

Early experience during model training showed that model performance was very sensitive to the learning rate. In order to negate these effects, we applied the learning rate reduction scheme \verb|ReduceLROnPlateau|, provided by the \verb|fastai| framework~\cite{Howard_2020}, to each model. The callback reduces the learning rate on each epoch if the training loss metrics are unchanging between consecutive epochs. This has given good results in studies such as \cite{alkababji2022scheduling} but at the cost of longer training times.

The computational environment for the experiments was provisioned on a virtual machine, equipped with 8 CPU cores (Intel Xeon Platinum 8260 CPU @ 2.40GHz) and 16GB of RAM. No hardware acceleration or GPUs were available.

%%%%%%%%%%%%%%%%%%%%%%%%%%%%%%%%%%%%%%
\subsection{Data}\label{lab:data}
%%%%%%%%%%%%%%%%%%%%%%%%%%%%%%%%%%%%%%
Due to commercial, legal and security considerations, there are very few well-labelled spacecraft anomaly datasets available to the public. The "SMAP/MSL" dataset provided by \cite{Hundman.2018} is a dataset used in other studies into autonomous detection of spacecraft anomalies (i.e. ~\cite{Baireddy.2021}, an LSTM-based study, and ~\cite{liu2023spacecraft}, a CNN-based approach). This dataset consists of curated telemetry streams from NASA's Soil Moisture Active Passive (SMAP)~\cite{smap} and Mars Science Laboratory "Curiosity rover" (MSL)~\cite{msl} missions. We will have selected this as the dataset for our study because it offers a good baseline against which to compare our results.

\begin{table}[!h]
    \centering
    \caption[SMAP/MSL Dataset Statistics]{\bf{SMAP/MSL Dataset Statistics, from \cite{Hundman.2018}}}
    \label{tab:smap_msl_stat}
    \begin{tabular}{l|r|r|r}
         &  \textbf{SMAP}&  \textbf{MSL}& \textbf{Total}\\ \hline
         Total anomaly sequences&  69&  36& 105\\
         Point anomalies&  43&  19 & 62\\
         Contextual anomalies&  26&  17& 43\\
         Unique telemetry channels&  55&  27& 82\\
         Input dimensions&  25&  55&  -\\
         Telemetry values evaluated&  429,735&  66,709& 496,444\\
    \end{tabular}
\end{table}

The data in \cite{Hundman.2018} has been scaled from between (-1,1) and anonymised. "\textit{Model input data also includes one-hot encoded information about commands that were sent or received by specific spacecraft modules in a given time window.}"~\cite{Hundman.2018}. This results in a collection of 82 multivariate data sets, with around 100 labelled anomalies in total across all data sets, as detailed in Table~\ref{tab:smap_msl_stat}. Each telemetry channel is a multivariate time series of one \textit{target} parameter and additional parameters to be used as contextual information. The target parameter is the time series to be forecast, in which anomalies are to be detected.

The data was pre-split by \cite{Hundman.2018} into "train" anomaly-free data to establish the nominal conditions and "test" sets, one per telemetry channel, which contain the labelled anomalies. We used the same split as in the original study in order to have comparable results.  

\subsection{Data Clustering}
Initial inspection of the telemetry channels show that different telemetry channels had varying general characteristics such as "spiky" or "flat". We wanted to investigate the link between the characteristics of the telemetry channels and the best performing deep learning model architecture, and whether specific architectures work better for certain types of data. Manual classification is not feasible due to the number of telemetry channels, so our idea was to use an unsupervised clustering approach.

\begin{figure}[H]
    \centering
    \includegraphics[width=0.65\linewidth]{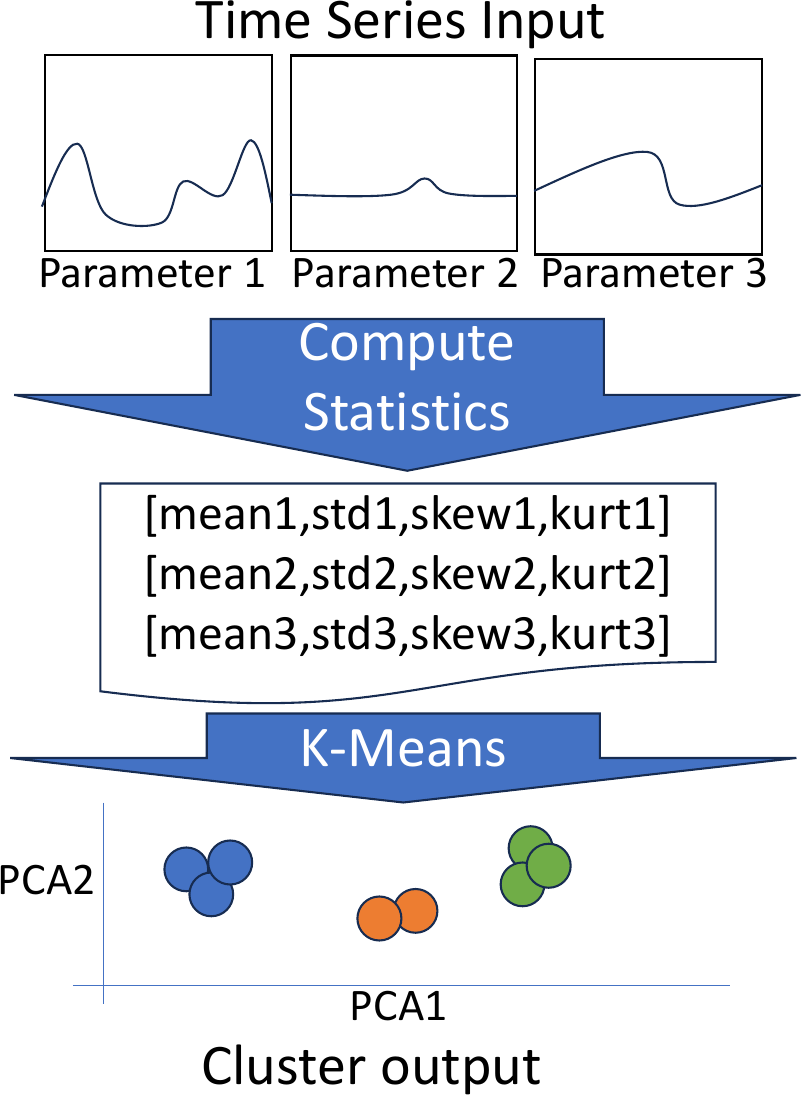}
    \caption{\bf{Clustering Process}}
    \label{fig:clustering_process}
\end{figure}

To associate the telemetry channels into clusters, we used an unsupervised clustering approach. Each class represents a particular set of characteristics. The method used the standard central moments (mean, standard deviation, skewness and kurtosis) calculated for the target parameter of each telemetry channel using \verb|SciPy|~\cite{Virtanen2020}. \verb|NaN|\footnote{Not a Number - used to signify an arithmetic error} values are set to \(0\). Therefore each telemetry channel was represented by a single four-dimensional vector. We applied K-Means clustering~\cite{IKOTUN2023178} to these four dimensional vectors, as illustrated in Figure~\ref{fig:clustering_process} and further elaborated in Listing~\ref{lst:cluster}.

\begin{figure*}[!ht]
    \centering
    \includegraphics[width=0.75\linewidth]{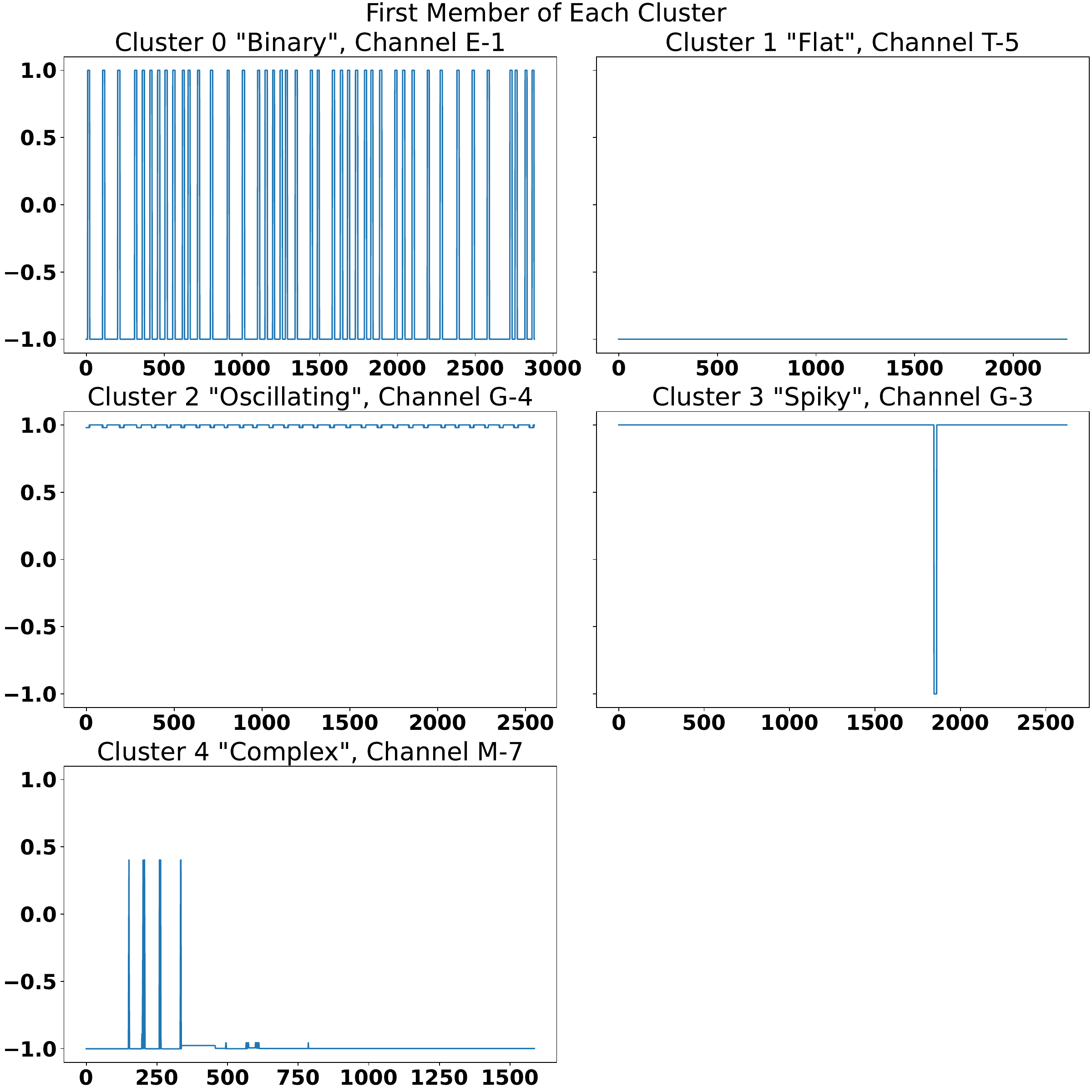}
    \caption{\bf{Data Types per Cluster}}
    \label{fig:channel_data_types}
\end{figure*}

The handling of \verb|NaN| values is required for the statistics skewness and kurtosis because some telemetry channels contain parameter values with no variance ("flat", in Table~\ref{fig:channel_data_types}); these are forced to zeros. Skewness measures the asymmetry of the probability distribution. For a constant data series, skewness is not defined, as skewness presupposes that there is variance in the data. Kurtosis measures the "tailedness" of the distribution. For a constant series, like skewness, kurtosis is also not defined because kurtosis measures the outliers and a constant series has none. Mean and standard deviation are defined in case of constant value so do not need to be treated for \verb|NaN|s. Skewness and Kurtosis are calculable for non-flat telemetry channels and give a better summary of the data than mean and standard deviation alone.

Our clustering focuses on the training data set, without anomalies, so as to identify what the "normal" behaviour of the parameter is, as summarised by the shape of the curve. In spacecraft operations this is the more likely scenario as often data channels have yet to experience an anomaly~\cite{Baireddy.2021},~\cite{Pilastre.2020}.

\begin{lstlisting}[label={lst:cluster},caption={Time Series Clustering Pseudo-Code}, float={ht},captionpos=b,frame=tb,aboveskip=-0em,belowskip=-4em]
For each telemetry channel i:
    Extract target parameter p_i from i
    Calculate central moments of p_i:
        [Mean, standard deviation,
         skew, kurtosis] => vector_i
    Set any value (vector_ij = NaN) => 0
    Add vector_i to list

Apply K-Means to list => n clusters
\end{lstlisting}

\begin{figure}[h]
    \centering
    \includegraphics[width=1\linewidth]{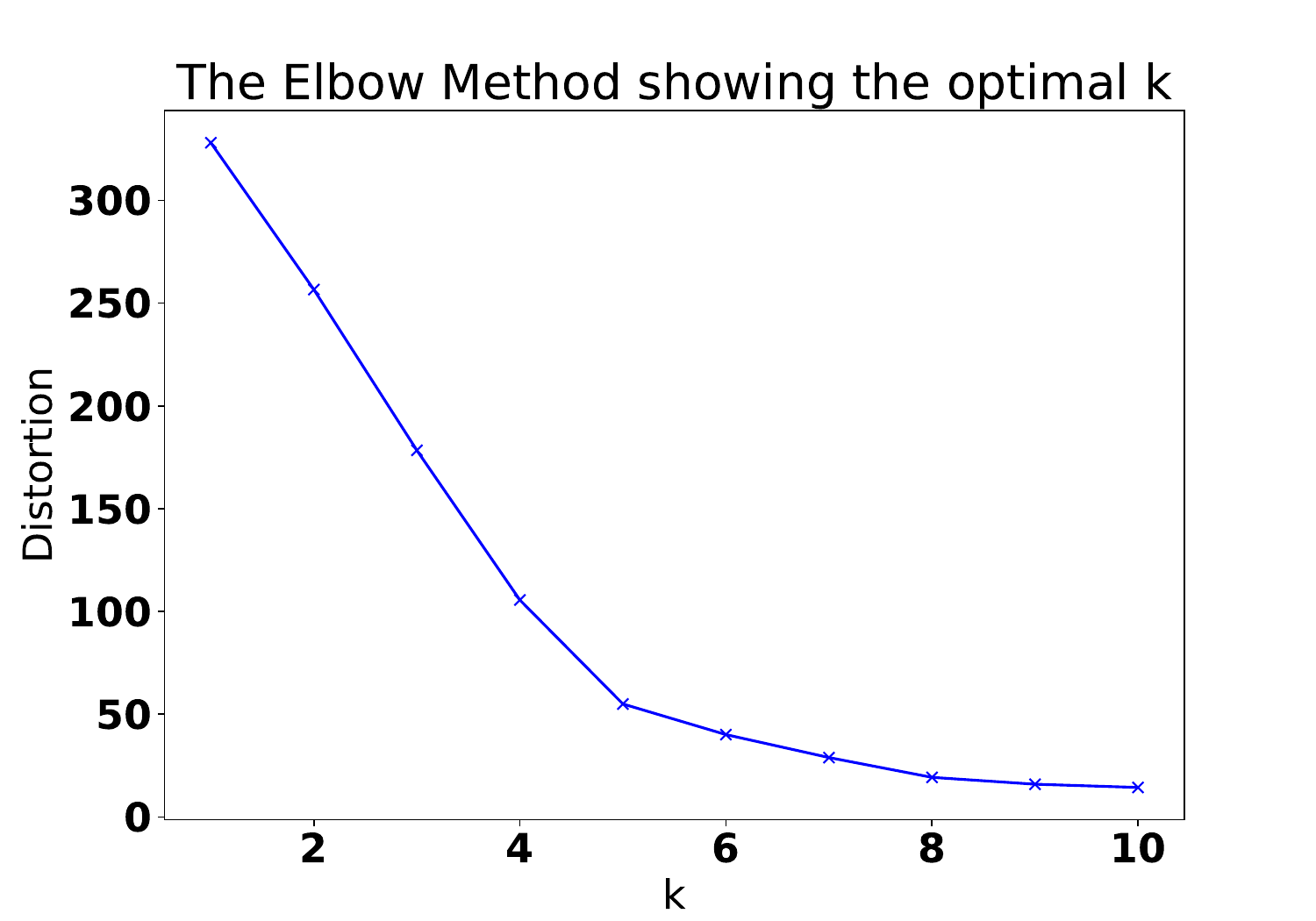}
    \caption{\bf{Elbow Method used to Determine Optimum Number of Clusters}}
    \label{fig:elbow}
\end{figure}

The "elbow method"~\cite{Marutho.2018} is a heuristic to find an optimal number of clusters by looking for a change in slope. For the \cite{Hundman.2018} data, the method indicated that 5 clusters of data types would be an optimal solution, as shown in Figure~\ref{fig:elbow}. The change in slope at \(k=5\) is clear. Distortion is an indication of how well the clusters fit, and \(k\) is the number of clusters. Lower values of \(k\) would suggest insufficiently separated clusters, whereas greater values would indicate overly split clusters. This result is dataset-specific, and may not reflect all spacecraft telemetry channels, however the K-means method is portable to other data sets and fast: all 82 channels were processed in under a second.

\begin{figure}[h]
    \centering
    \includegraphics[width=1.0\linewidth]{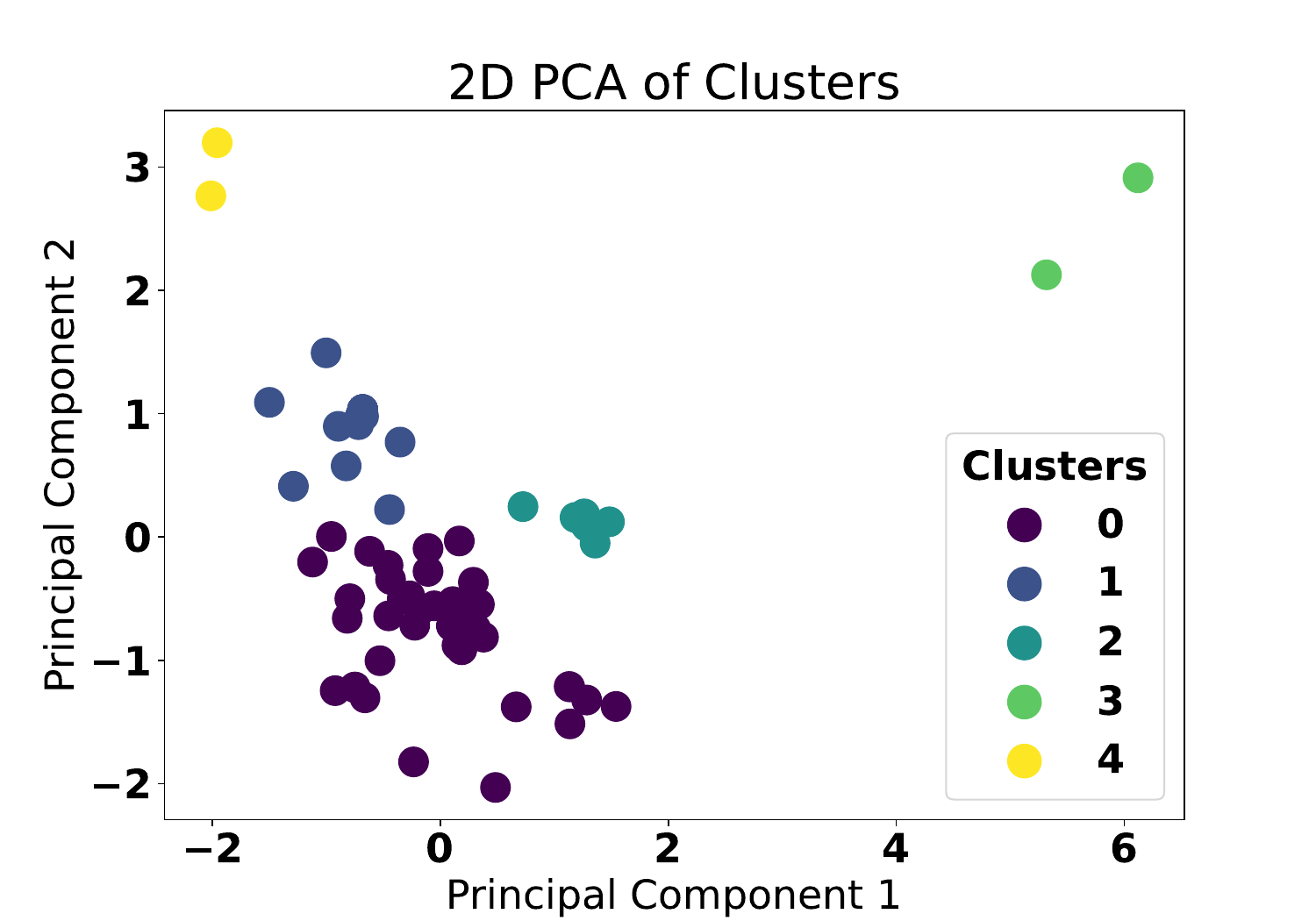}
    \caption{\bf{Resulting Data Clusters when k = 5}}
    \label{fig:clusters}
\end{figure}

The resulting clusters are shown in Figure~\ref{fig:clusters}. Each dot represents a telemetry channel target parameter, and the clusters are grouped by colour. Principle component analysis (PCA)~\cite{MACKIEWICZ1993303} has been used to reduce the number of dimension from 4 to 2 for the purposes of visualisation. Despite being few in number, the telemetry channels comprising clusters 3 ("Spiky") and 4 ("Complex") are clearly separated, with the remaining clusters being closer together yet still distinct.

The outcome of this investigation is illustrated in Figure~\ref{fig:channel_data_types}, and can be described as the following "types" of telemetry data according to the behaviour of the target data channel (that is, the one to be predicted by the forecasting model):
\begin{itemize}
    \item Cluster 0 "Binary": the values alternate between one of two values. When scaled to (-1,1), this shows as large spikes across the full range. There are 43 data channels in this cluster.
    \item Cluster 1 "Flat": the value is not expected to change at all. There are 21 data channels in this cluster.
    \item Cluster 2 "Oscillating": Similar to flag, but the value oscillates around a certain value rather than being fixed. There are 11 data channels in this cluster.
    \item Cluster 3 "Spiky": occasional large changes in the data are expected and normal. These represent a particular challenge for univariate models as the cause of the spike can only be determined from additional data. There are 2 data channels in this cluster.
    \item Cluster 4 "Complex": combination of the other data types. There are 2 data channels in this cluster.
\end{itemize}

In addition to reporting the results per model architecture trained on all telemetry channels, the best results per (model architecture, cluster) combination will also be given. This will inform whether certain architectures work better on certain types of data (cluster), or whether there is a "one-size-fits-all" universal solution which is applicable to all types of data behaviour.

%%%%%%%%%%%%%%%%%%%%%%%%%%%%%%%%%%%%%%
\section{Experiments and Results}
%%%%%%%%%%%%%%%%%%%%%%%%%%%%%%%%%%%%%%
The results of the 13 different models (described in Section~\ref{lab:archs}) shows considerable difference in training times, ranging from a few hours to over one full day, as shown in Table \ref{tab:results}. However the performance of the models do not scale with processing time (Table \ref{tab:results}). The performance is measured by two key metrics, the F1 (\%) score considering the number of anomalies correctly detected ("\textit{F1 anomaly}"), and the F1 score of the number of time points correctly labelled as occurring within anomalies ("\textit{F1 time point}").

Table \ref{tab:results} shows, per model architecture implementation, the total training time and the average training time per channel. True positive (TP), False positive (FP) and False negative (FN) values are also given per anomaly.

\begin{figure*}[h]
    \centering
    \includegraphics[width=0.9\linewidth]{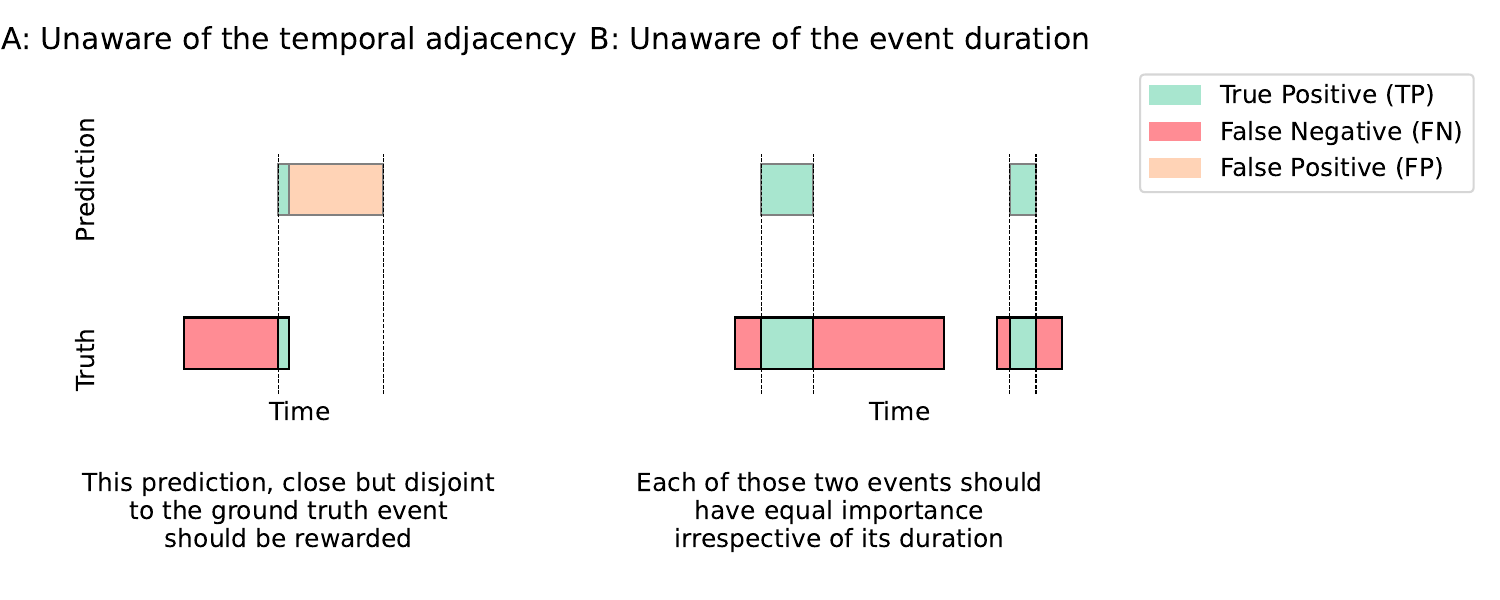}
    \caption{\bf{Anomaly detection metrics, per anomaly versus per time point}}
    \label{fig:range_metrics}
    \raggedright\footnotesize{Adapted from \cite{Huet2022} and \cite{tatbul.2018}}
\end{figure*}

It is expected that "\textit{F1 time point}" will not be very high, as the nature of the threshold-based anomaly detector means that data points either side of a labelled anomaly may not be detected as anomalous themselves, even though a domain expert would label them as such. Nevertheless, it gives an indication of the overall model performance when determining if any given data point is anomalous. This is illustrated in Figure~\ref{fig:range_metrics}, whereby a predicted anomaly and actual anomaly may share few actual data point yet nevertheless be considered a successful detection of an anomaly. That is, any overlap of predicted anomaly and actual anomaly is considered a detection, no matter how small (how few data points are correctly labelled). This is the metric used in \cite{Hundman.2018} and we retain it to allow direct comparison of results between their study and ours.

The F1 score pertaining to the detected anomalies ("\textit{F1 anomaly}") is more significant in terms of perceived anomaly detection performance by the spacecraft operator \cite{Hundman.2018,Baireddy.2021}.  

\subsection{Model Performance}
The best performing model architecture in our study is the CNN-based XceptionTimePlus implementation, with \textit{F1 anomaly} score  of \(69.9\%\). This is lower than the tuned results from the Telemanom study (Table~\ref{tab:telemanom_perf}) but represents a 6\% better performance than the worst performing model here, FCNPlus. It is noteworthy that the best and worst performing models are both of the CNN architecture families (XceptionTimePlus \(69.9\%\), FCNPlus \(63.2\%\)). This suggests that there is no intrinsic advantage of CNN-based models in general.

The hybrid TransformerLSTMPlus (\(69.6\%\)) and RNN-based GRUPlus (\(69.1\%\)) show similar performance although with vastly different training times.

Given the overall good performance of XceptionTimePlus (\(69.9\%\)) and the relatively low training time, this architecture would be our recommendation for a general purpose anomaly detector, as an initial investigation, before extensive effort it applied to tuning.

\subsection{Training Performance}
\begin{figure}[H]
    \centering
    \includegraphics[width=1\linewidth]{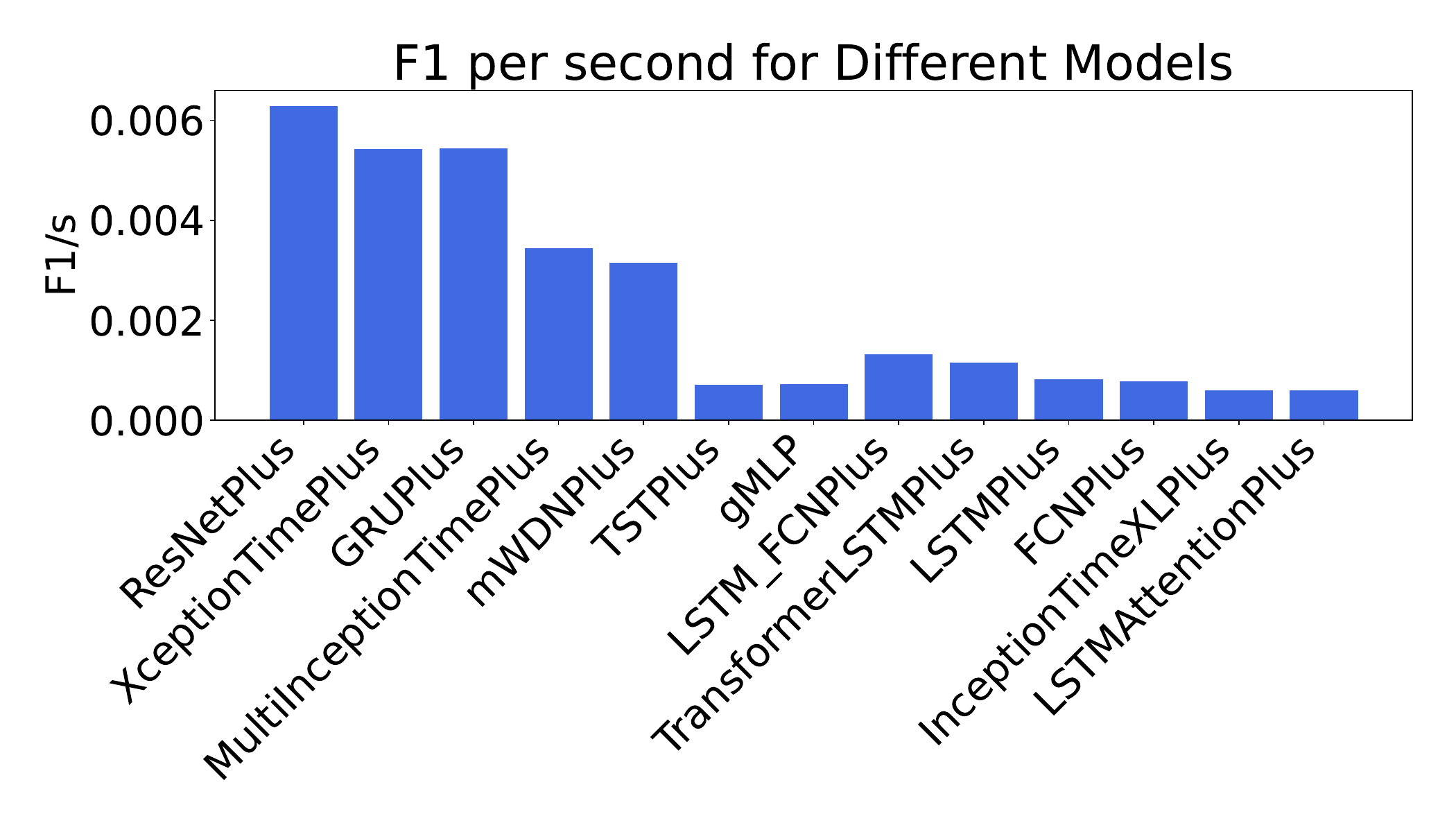}
    \caption{\bf{F1 score per training second}}
    \label{fig:f1_per_s}
\end{figure}

Spacecraft parameters number in the tens to hundreds of thousands per spacecraft, so the effectiveness of the models in terms of training is of critical importance. Training a large number of slow-to-train models is thus potentially prohibitive, and models would need regular retraining as the spacecraft ages or environment changes. We introduce a 'F1 per second' (F1/s) metric, to provide a comprehensive view of our models' performance, taking into account both their accuracy and computational efficiency. In essence, it indicates the F1 score achieved for every second of training. A higher value implies that the model not only performs well but can be trained quickly, making it both effective and efficient. Figure~\ref{fig:f1_per_s} shows that the CNN-based models ResNetPlus and XceptionTimePlus are the most efficient, whereas the "large" (many model parameters) CNN-based InceptionTimeXLPlus and RNN/Transformer hybrid LSTMAttentionPlus offer the worst performance/training time trade-off. Generally the Transformer-based architectures all suffer from a performance/training time trade-off. The difference in training times is stark - from 2.5 minutes for XceptionTimePlus to over 18.5 minutes for TransformerLSTMPlus, whereas their F1 score is nearly identical (\(69.9\%\) vs \(69.6\%\)). 

\begin{table*}[!ht]
    \caption{\bf{Results per Architecture}}
    \label{tab:results}
    \centering
    \begin{tabular}
    {l|p{18mm}p{15mm}|p{14mm}|p{13mm}p{13mm}p{13mm}p{13mm}}
    \hline
        \textbf{Architecture} & \textbf{Total Time} \newline \footnotesize(days hh:mm:ss)&\textbf{ Avg Time per TM Channel} \footnotesize(mm:ss)& \textbf{F1 (\%)} Time~Point& \textbf{TP Anomaly}& \textbf{FP Anomaly}& \textbf{FN Anomaly}& \textbf{F1 (\%)} Anomaly\\ \hline
  \textbf{XceptionTimePlus}
& 00 03:34:27
& 02:39
& 37.42
& 65
& 16
& 40
& \textbf{69.89}
\\
 TransformerLSTMPlus
& 01 00:58:58
& 18:30
& 36.69
& 64
& 15
& 41
&69.57
\\ 
        GRUPlus
& 00 03:32:16
& 02:37
& 32.71
& 66
& 20
& 39
& 69.11
\\ 
        ResNetPlus
& \textbf{00 03:02:39}
& \textbf{02:15}
& \textbf{38.36}
& 63
& 15
& 42
& 68.85
\\ 
        MultiInceptionTimePlus
& 00 05:32:11
& 04:06
& 36.71
& 60
& \textbf{10}
& 45
& 68.57
\\ 
        LSTM\_FCNPlus
& 00 23:15:57
& 17:14
& 32.83
& 66
& 23
& 39
& 68.04
\\ 
        LSTMPlus
& 01 02:21:02
& 09:46
& 35.10
& \textbf{67}
& 25
& \textbf{38}
& 68.02
\\ 
        mWDNPlus
& 00 05:53:12
& 04:22
& 34.99
& 63
& 21
& 42
& 66.67
\\ 
        gMLP
& 00 16:06:24
& 11:56
& 32.44
& 63
& 21
& 42
& 66.67
\\
 TSTPlus
& 00 13:29:22
& 10:00
& 34.94
& 57
& 16
& 48
& 64.04
\\
 InceptionTimeXLPlus
& 01 05:38:24
& 21:57
& 33.32
& 64
& 31
& 41
& 64.00
\\
 LSTMAttentionPlus
& 01 05:41:37
& 22:00
& 35.35
& 62
& 29
& 43
& 63.27
\\
 FCNPlus
& 01 00:57:41
& 09:15
& 36.29
& 61
& 27
& 44
& 63.21
\\ 
    \end{tabular}
\end{table*}

\subsection{Best Architecture per Cluster}
\begin{table*}[ht!]
    \caption{\bf{Results per Cluster}}
    \label{tab:results_per_data_type}
    \centering
    \begin{tabular}%{l|l|l{14mm}p{14mm}}
    {l|l|r|r}
    \hline
        \textbf{Architecture} & \textbf{Cluster} & \textbf{F1 (\%)} Time Point& \textbf{F1 (\%)} Anomaly\\ \hline
\textbf{MultiInceptionTimePlus}& Binary& 31.65& \textbf{65.22}
\\
 ResNetPlus& Binary& \textbf{34.02}&64.58
\\ \hline
        \textbf{gMLP}& Complex& \textbf{43.2}4& \textbf{100.00}
\\
 LSTM\_FCNPlus
& Complex& 42.87&100.00 
\\ \hline
        \textbf{XceptionTimePlus}& Flat& \textbf{44.10}& \textbf{86.27}
\\
 TransformerLSTMPlus& Flat& 41.00&85.19
\\ \hline
        \textbf{XceptionTimePlus}& Oscillating& \textbf{27.50}& \textbf{72.00}
\\
 ResNetPlus& Oscillating& 26.87&64.00
\\ \hline
        \textbf{gMLP}& Spiky& 48.87& \textbf{100.00}
\\
 MultiInceptionTimePlus& Spiky& 35.08&85.71
\\ \hline 
 \textit{Average} & - & - & \textit{84.7}
\\
    \end{tabular}
\end{table*}

A further level of analysis into the results provides insight into the best performing models per type of telemetry data, as determined by the clusters shown in Figure~\ref{fig:channel_data_types}. To calculate the \textit{F1 anomaly} scores per time point and per labelled anomaly, the true positives, false positives and false negatives (in each case) were summed across the relevant cluster.

All (architecture, cluster) pairs were ranked by \textit{F1 anomaly} score, to determine the best performing model for each cluster. The architectures identified as best and second-best performing for each cluster are given in Table~\ref{tab:results_per_data_type}. 

It is notable that the best performing architecture identified in Table~\ref{tab:results} (XceptionTimePlus, \(69.9\%\)) is not the best in each data type cluster. The CNN-based MultiInceptionTimePlus and XceptionTimePlus models perform best in the "Binary" (\(65.2\%\)) and "Flat"/"Oscillating" clusters (\(86.3\%\), \(72.0\%\)) respectively, suggesting that the spatial awareness of the models is particularly useful in identifying anomalies in these cases. 

With fewer telemetry channels, two apiece, the Spiky and Complex clusters are more difficult to assess in general terms due to the low number of examples in the data set. Despite being an older architecture, the MLP-based gMLP performs best for the "Spiky" (\(100\%\)) and "Complex" cases (\(100\%\)), outperforming all other architectures. More examples of these clusters are needed before a recommendation can be made on these specific telemetry data types, but it is instructive to note the differences in performance. With an average F1 score of \(84.7\%\), the best performing models per cluster collectively outperform any single model by nearly \(15\%\) (absolute). 

As an ensemble approach to the general anomaly detection problem, taking the best performing architecture per-data type greatly increases the performance overall. The average F1 score of \(84.7\%\) surpasses the \(83.6\%\) achieved by the baseline study, Telemanom~\cite{Hundman.2018}.

%%%%%%%%%%%%%%%%%%%%%%%%%%%%%%%%%%%%%%
\section{Conclusion and Future Work}
%%%%%%%%%%%%%%%%%%%%%%%%%%%%%%%%%%%%%%
In conclusion, the insights derived from our study have shown innovative advancements in spacecraft anomaly detection, laying a robust foundation for future explorations and discoveries in this domain.

\subsection{Conclusion}
In this work, we performed a comparative study of diverse deep learning model architectures, with the goal of assessing their efficacy in spacecraft anomaly detection. Our findings revealed that model XceptionTimePlus (\(69.9\%\)) exhibited the most optimal performance among all the models assessed in the study, across all telemetry channels. However, it is important to note that the overall performance was not on par with the outcomes demonstrated in~\cite{Hundman.2018}. A contributing factor to this is the conscious decision to refrain from hyperparameter optimisation in order to preserve the default comparisons and allow direct relative comparisons. Nevertheless, our study provides valuable insights into which families of deep learning architecture perform well, and which not. 

Furthermore, due to constraints in computational resources it was not possible to follow the standard optimisation strategies such as grid search, which runs many iterations of the model to explore the hyperparameter space. With some models taking several days to run once (e.g. LSTMAttentionPlus at 1 day and 5 hours), it is infeasible to run the large number of iterations required.

In addition to this, our research illuminated that different deep learning model architectures exhibit varying degrees of proficiency depending on the nature of the data, be it "spiky", "flat", "complex", "oscillating", or "binary". We introduced an innovative clustering methodology in this paper, facilitating the efficient allocation of spacecraft telemetry channels into distinct clusters contingent on the inherent statistical properties of the data, based on the shape of the curve. This novel approach has not only advanced our understanding but has also paved the way for the advent of more sophisticated ensemble models, based on individual models that are harmoniously optimized for disparate data types. This ensemble approach was able to exceed the performance of the baseline study (\(84.7\%\) vs \(83.6\%\)), despite using unoptimised models.

\subsection{Future Work}
The work in our study has suggested new possibilities and directions for future research. A natural extension of this work would be the exploration of ensemble models that are proficiently optimised to accommodate various data types, leveraging the clustering methodology introduced in this paper. Furthermore, a meticulous exploration of hyperparameter space will be pivotal to harness the maximal potential of the models, thereby advancing the state-of-the-art in spacecraft anomaly detection. 

As described above, the individual models were not individually optimised per model, rather used defaults from the respective frameworks \verb|fastai| and \verb|tsai|. The success of the clustering approach suggests itself as an alternative approach to the one-model-for-all approach seen in other studies (\cite{Baireddy.2021,Hundman.2018,Heras.2014}): that of creating a \textit{set} of optimised hyperparameters per data type (spiky, binary, etc). 

Additionally, current anomaly detection approaches (e.g. \cite{Heras.2014,Hundman.2018,Pilastre.2020,Baireddy.2021}) rely predominantly on forecasting models to deduce nominal behavior, identifying anomalies through a comparative analysis of predictions against predetermined thresholds. A promising avenue for future research would be the application of deep learning classification techniques, which could potentially offer a direct assessment of the telemetry channels without relying on thresholds.

\FloatBarrier

%%%%%%%%%%%%%%%%%%%%%%%%%%%%%%%%%%%%%%%%%%%%%%%%%%%%%%%%%%%%%%%%%%%%%%%%%%%%%%%%%%%%%%%%%%%%%%%%%%%%%%
\bibliographystyle{IEEEtran}
\bibliography{biblio}

\FloatBarrier
%%%%%%%%%%%%%%%%%%%%%%%%%%%%%%%%%%%%%%%%%%%%%%%%%%%%%%%%%%%%%%%%%%%%%%%%%%%%%%%%%%%%%%%%%%%%%%%%%%%%%%
\thebiography
\begin{biographywithpic}
{Daniel Lakey}{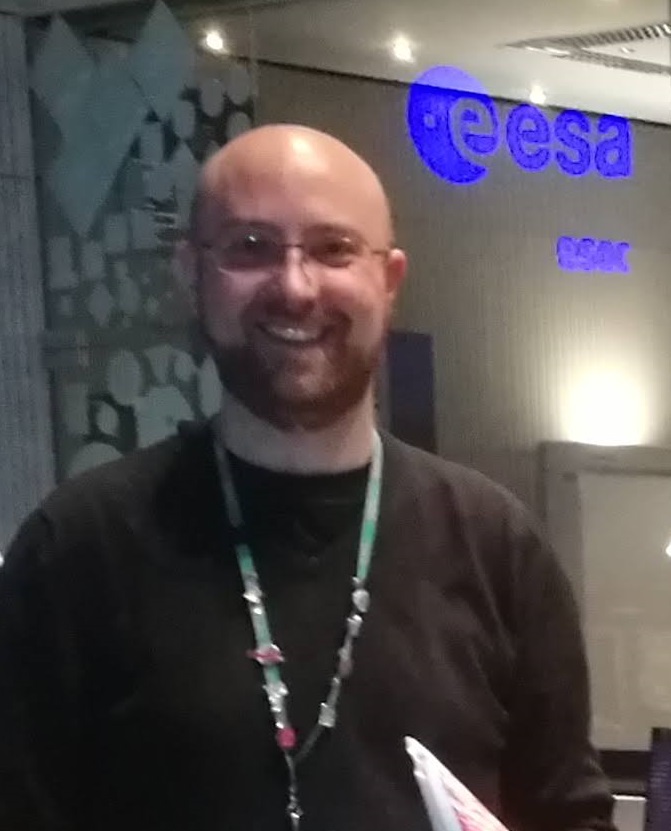}
received his BSc degree in Computer Science in 2003 from Cardiff University, and is completing a MSc in Data Science with IU International University of Applied Sciences. Working with the European Space Agency, Daniel has been deeply involved with interplanetary exploration missions since 2006. Since 2013 Daniel has been a Spacecraft Operations Engineer on the ESA/NASA Solar Orbiter mission, with a particular focus on anomaly investigation and resolution.\\
\href{https://orcid.org/0000-0002-8198-7892}{orcid.org/0000-0002-8198-7892}
\end{biographywithpic} 

\begin{biographywithpic}
{Tim Schlippe}{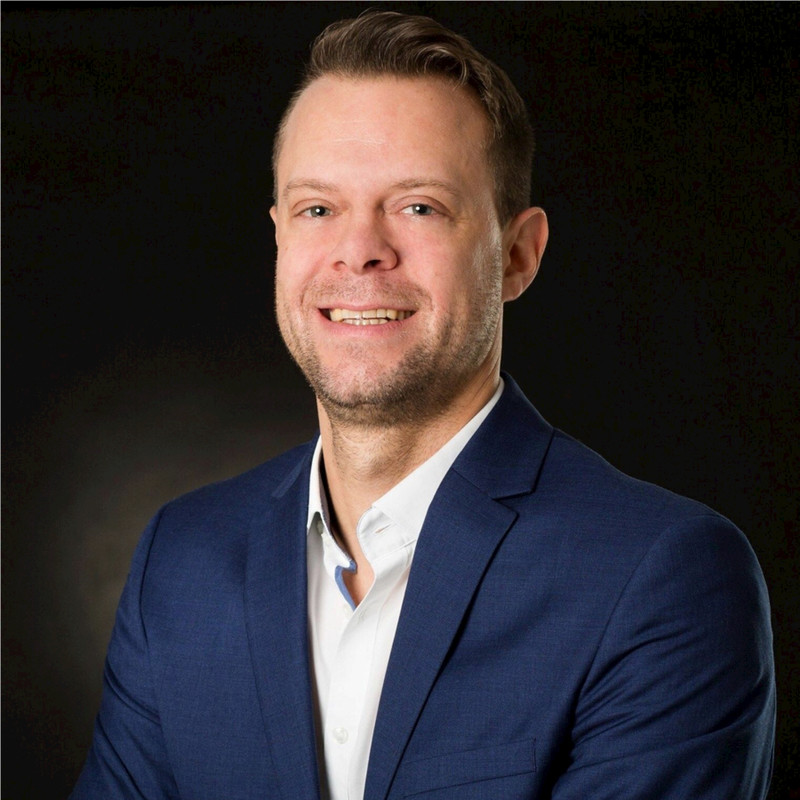}
is a professor of Artificial Intelligence at IU International University of Applied Sciences and CEO of the company Silicon Surfer. Prof. Dr. Schlippe has in-depth knowledge in the fields of artificial intelligence, machine learning, natural language processing, multilingual speech recognition/synthesis, machine translation, language modeling, computer-aided translation, and entrepreneurship, which can be seen in his numerous publications at international conferences in these areas. \\
\href{https://orcid.org/0000-0002-9462-8610}{orcid.org/0000-0002-9462-8610}

\end{biographywithpic}

\end{document}